\documentclass[letterpaper, 10 pt, conference]{ieeeconf}  % Comment this line out if you need a4paper

\IEEEoverridecommandlockouts                              % This command is only needed if thanks

\newtheorem{definition}{Definition} 
\newtheorem{theorem}{Theorem}
\newtheorem{problem}{Problem}

\usepackage{graphics} % for pdf, bitmapped graphics files
\usepackage{hyperref}

\title{\LARGE \bf
Model Predictive Path-Following for Constrained \\ Differentially Flat Systems
}

\author{Melissa Greeff and Angela P. Schoellig
\thanks{The authors are with the Dynamic Systems Lab ({www.dynsyslab.org}) at the University of Toronto Institute for Aerospace Studies (UTIAS), Canada. melissa.greeff@mail.utoronto.ca, schoellig@utias.utoronto.ca}%
\thanks{This work was supported by Drone Delivery Canada, Defence Research and Development Canada and Natural Sciences and Engineering Research Council of Canada.}}%

\begin{document}

\maketitle
\thispagestyle{empty}
\pagestyle{empty}

%%%%%%%%%%%%%%%%%%%%%%%%%%%%%%%%%%%%%%%%%%%%%%%%%%%%%%%%%%%%%%%%%%%%%%%%%%%%%%%%
\begin{abstract}

For many tasks, predictive path-following control can significantly improve the performance and robustness of autonomous robots over traditional trajectory tracking control. It does this by prioritizing closeness to the path over timed progress along the path and by looking ahead to account for changes in the path. We propose a novel predictive path-following approach that couples feedforward linearization with path-based model predictive control. Our approach has a few key advantages. By utilizing the differential flatness property, we reduce the path-based model predictive control problem from a nonlinear to a convex optimization problem. Robustness to disturbances is achieved by a dynamic path reference, which adjusts its speed based on the robot's progress. We also account for key system constraints. We demonstrate these advantages in experiment on a quadrotor. We show improved performance over a baseline trajectory tracking controller by keeping the quadrotor closer to the desired path under nominal conditions, with an initial offset and under a wind disturbance. 

\end{abstract}

%%%%%%%%%%%%%%%%%%%%%%%%%%%%%%%%%%%%%%%%%%%%%%%%%%%%%%%%%%%%%%%%%%%%%%%%%%%%%%%%
\section{INTRODUCTION}

% specific examples 
As autonomous robots, such as unmanned aerial vehicles (UAVs), wheeled vehicles, legged robots and mobile manipulators, are more substantially utilized in tasks including manufacturing, transportation, mapping \cite{c1}, and inspection \cite{c2}, they are required, even in the event of disturbances, to exhibit high-performance and safe behavior. 

One potentially successful approach to ensure this safe behaviour is to enhance \textit{robustness} to disturbances by prioritizing closeness to a desired geometric path. In path-following, in contrast to trajectory tracking, the reference is a geometric path exempt from any preassigned time-dependency. Path-following overcomes a limitation of trajectory tracking which is that, if the robot loses track of the reference, it may discourse from the path in an attempt to align with the time-dependent reference and as such potentially collide with surrounding obstacles, reach saturation limits or exhibit some other unsatisfactory behaviour \cite{c3}.

Another potentially successful approach to ensure this safe behaviour is to enhance \textit{performance} by predicting ahead in order to prepare for changes in the path.  Model predictive control (MPC) has demonstrated high-performance \cite{c4} by optimizing over a prediction horizon while still explicitly adhering to constraints on the states and inputs of the system \cite{c5}.

   \begin{figure}[thpb]
      \centering
      \includegraphics[scale=1.0]{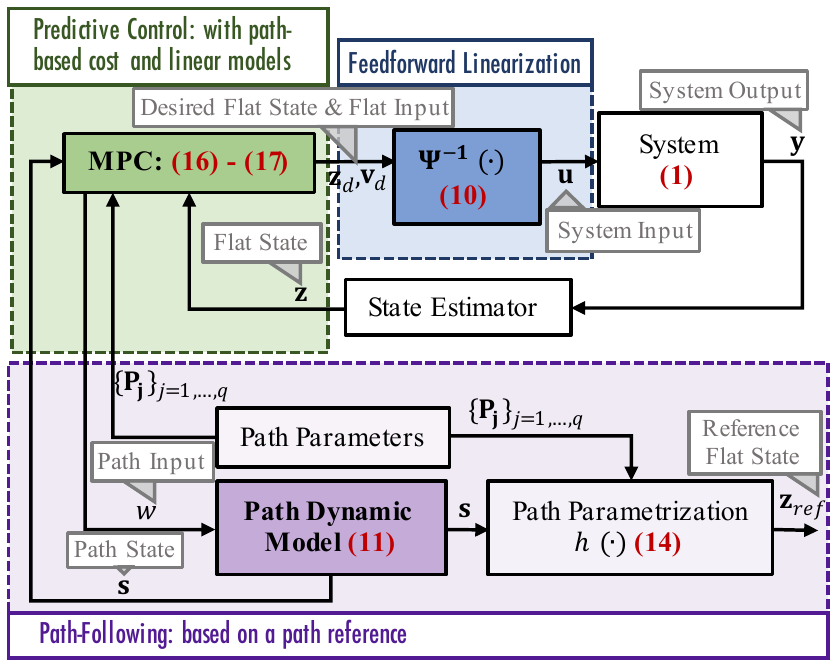}
      \caption{Overall architecture diagram of the proposed Flatness Approach to Predictive Path-following (FAPP).}
      \label{frapp}
      \vspace{-5mm}
   \end{figure}

Considering these approaches, the aim of this paper is to design and implement a controller that predicts ahead to accurately and robustly follow a curved path while adhering to key system constraints. In particular, we consider predictive control with a path-based cost, which, when making use of the differential flatness property of many robotic systems, is subject to only linear models as demonstrated in Fig. 1. In order to do this, we now consider literature in three areas: path-following, MPC and differentially flat systems. 

Many path-following approaches are not easily integrated into the optimal control problem (OCP) of a model predictive framework because their objective is to modify existing trajectory tracking control or to stabilize a set of trajectories. One such approach, from the early work on path-following, considers converting an existing tracking controller into a path-following controller by re-weighting the relative importance of spatial convergence to time convergence \cite{c6}. Another more recent approach considers transverse feedback linearization to stabilize the path-following manifold or the set of trajectories that allow convergence to the path \cite{c7}. Consideration of state and input constraints is difficult in both approaches. 

Two path-following approaches have been used in a model predictive framework. In the first approach \cite{c8}, the kinematic model of a unicycle vehicle is rewritten in terms of a spatial and orientation error model defined with respect to a path-attached frame. The rotation rate of the vehicle is controlled using static feedback linearization of this error model. This approach has been extended to work within Nonlinear Model Predictive Control (NMPC) \cite{c9} for visual teach and repeat (VT\&R) \cite{c10}. However, to avoid a singularity in the error model, the vehicle needs to be closer to the path than a distance of one over the maximum path curvature. 

To avoid this restriction, we consider an alternative approach where a separate path dynamic is proposed \cite{c3}. Specifically, the desired geometric path is parametrized in terms of a path variable $\theta$ whose dynamics describe the motion of a virtual vehicle along the path. Moreover, the path-following problem is separated into a geometric task, where the system is required to converge to the path, and a secondary dynamic task, where the system is required to satisfy a dynamic specification, such as a speed assignment, along the path. This path-following approach is combined with NMPC by extending the nonlinear system dynamics with the introduced path dynamic \cite{c11}, \cite{c12}. Fundamentally, in the NMPC the evolution of the path state, and correspondingly the reference on the path, is optimized alongside the system state. This describes the predictive path-following problem. 

A drawback of this predictive path-following problem, highlighted in simulation for a crane in \cite{c13} and quadrotor in \cite{c14}, is that the resulting optimal control problem is subject to two nonlinear models and system constraints making efficient implementation online in a high-frequency feedback control loop difficult. 

We investigate differential flatness, which in conjunction with linearization techniques, can be used to reduce the complexity of the control design. Many physical systems, including cranes, cars with trailers and quadrotors, can be described by nonlinear models exhibiting the property known as differential flatness \cite{c15}, \cite{c16}. Intuitively, differential flatness allows us to separate the nonlinear model into a linear dynamics component and a nonlinear transformation. 

This property has been utilized in both feedback and feedforward linearization  approaches wherein a linear controller is designed based on the linear dynamics term and an inverse term is then used to correct for the nonlinearity in the model \cite{c17}, \cite{c18}. Feedforward linearization differs fundamentally from feedback linearization in that the desired flat state instead of the measured flat state of the system is used in the inverse term. Feedforward linearization aims to overcome the robustness issue of feedback linearization. Specifically, this robustness issue is a result of a potential parametric uncertainty of the model leading to inexact pole-zero cancellation \cite{c19}. Given the computational and robustness advantages of feedforward linearization, our approach incorporates this idea to solve the predictive path-following problem. 

The contributions of this paper are then three-fold. Our first contribution is to introduce a Flatness Approach to Predictive Path-following (FAPP) which provides novelty in two aspects. Firstly, in contrast to related approaches in Section \ref{sect_ff_lin_pred}, it combines MPC and feedforward linearization in a novel way. Secondly, it provides novelty in demonstrating that by incorporating a key result from exact path-following in Section \ref{sect_ff_lin_pf}, we can reduce the predictive path-following problem to solving a quadratic program (QP) at each iteration. The second contribution is to detail how such an approach can be applied to a quadrotor in Section \ref{sect_quad}. In the final contribution, we demonstrate the value of FAPP in terms of robustness and accuracy over a conventional trajectory tracking controller on a quadrotor.

\section{PROBLEM STATEMENT}
\label{sect_problem}

Consider a continuous-time, nonlinear system model of the form:
\vspace{-4mm}
$$
\mathbf{\dot{x}}(t) = f(\mathbf{x}(t), \mathbf{u}(t)), \quad \mathbf{x}(0) = \mathbf{x_0}, \eqno{(1)}
$$
\noindent
with $t \in \mathbb{R}$, $ \mathbf{x}(t) \in X \subseteq \mathbb{R}^n$, $ \mathbf{u}(t) \in U \subseteq \mathbb{R}^m$ and $f$ a smooth function. Given a path $P$, described by a map that projects the real interval $[\theta_0, \theta_1]$ to the state space,
\vspace{-2mm}
$$ P = \{ \mathbf{x}_{ref} \in \mathbb{R}^n  | \quad \mathbf{x}_{ref} = \mathbf{p}(\theta(t)), \quad \theta \in [\theta_0, \theta_1] \}, \eqno{(2)}$$
we consider the path-following problem given in Problem 1.

\begin{problem}
\normalfont
	Design a controller that achieves the following:
	
	\textbf{P1 (Geometric Convergence)} The path error vanishes asymptotically:
	\vspace{-4mm}
	$$\lim_{t\to\infty} (\mathbf{x}(t) - \mathbf{p}(\theta(t))) = 0.$$
	
	\textbf{P2 (Dynamic Specification)} The system (1) achieves some dynamic specification, such as a desired speed, that moves it forward along the path.
	
	\textbf{P3 (Constraint Satisfaction)} The state constraints, $ \mathbf{x}(t) \in X \subseteq \mathbb{R}^n$, and input constraints, $ \mathbf{u}(t) \in U \subseteq \mathbb{R}^m$,  are adhered to for all time.
	
\end{problem}

Subsequently, we assume that (1) is \textit{differentially flat}, see Section \ref{sect_defn_diff_flat},  and propose a scheme for model predictive path-following control to solve the considered problem. We treat the path  parameter $\theta$ as an additional state variable in an expanded predictive setup and obtain its  evolution and the real system input by solving an open-loop optimal control problem. %Because of the differential flatness property, the predictive path-following problem can be  converted into a quadratic program that can be solved efficiently

\section{BACKGROUND}

\subsection{Differential Flatness}

\label{sect_defn_diff_flat}
We recall the formal definition of differential flatness. 

\noindent
\begin{definition}

\normalfont
A nonlinear system model (1) is \textit{differentially flat} if there exists $ \pmb{\zeta}(t) \in \mathbb{R}^m$, whose components are differentially independent, such that the following holds \cite{c16}:

\vspace{-4mm}
$$
\pmb{\zeta} = \Lambda(\mathbf{x}, \mathbf{u}, \mathbf{\dot{u}}, \hdots, \mathbf{u}^{(\delta)}) \eqno{(3)}
$$

\vspace{-7mm}
$$
\mathbf{x} = \Phi(\pmb{\zeta}, \pmb{\dot{\zeta}}, \hdots, \pmb{\zeta}^{(\rho-1)}) \eqno{(4)}
$$

\vspace{-7mm}
$$
\mathbf{u} = \Psi^{-1}(\pmb{\zeta}, \pmb{\dot{\zeta}}, \hdots, \pmb{\zeta}^{(\rho)}), \eqno{(5)}
$$

\noindent
where $\Lambda$, $\Phi$ and $\Psi^{-1}$ are smooth functions, $\delta$ and $\rho$ are the maximum orders of the derivatives of $\mathbf{u}$ and $\pmb{\zeta}$ needed to describe the system and $ \pmb{\zeta} = [\zeta_1, \hdots, \zeta_m]^T $ is called the flat output.
\end{definition}

\subsection{Feedforward Linearization}
\label{sect_key_ff_lin}

We briefly highlight a key result in feedforward linearization that demonstrates how we can rewrite the nonlinear model (1) as an equivalent linear one. As explained in \cite{c17}, every \textit{differentially flat} system (1) can be represented by a Brunovsk\'y state (or \textit{flat state}): 

\vspace{-4mm}
$$
\mathbf{z} := \begin{bmatrix} \zeta_1,  \dot{\zeta}_1, \hdots, \zeta_1^{(\rho_1-1)}, \hdots, \zeta_m, \hdots, \zeta_m^{(\rho_m-1)} \end{bmatrix}^T. \eqno{(6)}
$$

\noindent
Note that $\rho_i$ is the maximum derivative of $\zeta_i$ found in Ò(5)Ó. Using the state-transformation between the flat state $\mathbf{z}$ and state $\mathbf{x}$, obtained by differentiation of (3) and using (4), we can transform (1) into the normal form:

\vspace{-4mm}
$$
\zeta_i^{(\rho_i)} = \alpha_i(\pmb{\zeta}, \pmb{\dot{\zeta}}, \hdots, \pmb{\zeta}^{(\rho-1)}, \mathbf{u}, \mathbf{\dot{u}}, \hdots, \mathbf{u}^{(\sigma_i)}) := v_i,
\eqno{(7)}
$$

\noindent
where $\alpha_i$, $ i = 1 \hdots m$, is a smooth function obtained as a result of the transformation. Note $\sigma_i$ is the maximum derivative of $\mathbf{u}$ after $\rho_i$ times differentiating $\zeta_i$ in (3). We define the \textit{flat input} $\mathbf{v}$ as:

\vspace{-3mm}
$$
\mathbf{v} := \begin{bmatrix} v_1, v_2, \hdots, v_m \end{bmatrix}^T. \eqno{(8)}
$$

%if system Ò(1)Ó is differentially flat then it can be transformed using Ò(4)Ó into the normal form:

%\vspace{-2mm}
%$$
%\zeta_i^{(\rho_i)} = \alpha_i(\pmb{\zeta}, \pmb{\dot{\zeta}}, \hdots, \pmb{\zeta}^{(\rho-1)}, \mathbf{u}, \mathbf{\dot{u}}, \hdots, \mathbf{u}^{(\sigma_i)}) := v_i
%\eqno{(7)}
%$$

%\noindent
%for $ i = 1 \hdots m$. Note that $\rho_i$ is the maximum derivative of $\zeta_i$ found in Ò(6)Ó and $\sigma_i$ is the maximum derivative of $\mathbf{u}$ after differentiation $\rho_i$ times of $\zeta_i$ in (4).  We will define the \textit{flat state}, $\mathbf{z}$, and \textit{flat input}, $\mathbf{v}$ as:
%$$
%\mathbf{z} := \begin{bmatrix} \zeta_1,  \dot{\zeta}_1, \hdots, \zeta_1^{(\rho_1-1)}, \hdots, \zeta_m, \hdots, \zeta_m^{(\rho_m-1)} \end{bmatrix}^T, \eqno{(8a)}
%$$

%\vspace{-3mm}
%$$
%\mathbf{v} := \begin{bmatrix} v_1, v_2, \hdots, v_m \end{bmatrix}^T. \eqno{(8b)}
%$$

\noindent
Using the definitions in Ò(6)Ó and Ò(8)Ó, we rewrite Ò(7)Ó as:
\vspace{-1mm}
$$
\mathbf{\dot{z}} = \mathbf{A} \mathbf{z} + \mathbf{B} \mathbf{v}
\eqno{(9a)}
$$
\vspace{-4mm}
$$
\mathbf{v} = \Psi(\mathbf{z}, \mathbf{u},  \mathbf{\dot{u}}, \hdots, \mathbf{u}^{(\sigma)}),
\eqno{(9b)}
$$

\noindent
where $\sigma = \max{\sigma_i}$. We term (9a) the \textit{linear flat model}. By substituting the definitions in Ò(6)Ó and Ò(8)Ó, we can rewrite Ò(5)Ó as $ \mathbf{u} = \Psi^{-1}(\mathbf{z}, \mathbf{v})$.

\begin{theorem}

\normalfont
From \cite{c17}: Consider a desired trajectory in the flat output $\pmb{\zeta}_d$, including a corresponding desired flat state $\mathbf{z}_d$ (obtained by substituting $\pmb{\zeta}_d$ for $\pmb{\zeta}$ in Ò(6)Ó) and desired flat input $\mathbf{v}_d$ (obtained by substituting $\pmb{\zeta}_d$ for $\pmb{\zeta}$ in Ò(7)Ó and (8)). Given $\pmb{\zeta}_d$, if we apply the nominal control:
\vspace{-1mm}
$$ \mathbf{u} = \Psi^{-1}(\mathbf{z}_d, \mathbf{v}_d), \eqno{(10)}$$

\noindent
to a differentially flat system Ò(1)Ó, provided that $\mathbf{z}(0) = \mathbf{z}_d(0)$, this results in an equivalent, by change of coordinates, linear system as given in Ò(9a)Ó.
	
\end{theorem} 

Theorem 1 allows trajectory generators, as in Fig. 2, or controllers, as in our proposed approach in Fig. 1, to only consider the equivalent linear flat model. The output of the trajectory generator or controller, i.e. the desired flat state and flat input, can then be fed through the inverse transformation (10) to correct for the nonlinear part (9b) in the system. 

\subsection{Exact Path-Following}
\label{sect_ff_lin_pf}

We emphasize a key result from exact path-following (which is a subset of path-following problems that assumes an exact model of the system that starts precisely on the path). In \cite{c20}, the feedforward linearization is used to determine an optimal feedforward control that moves a differentially flat system forward along a path.   Sufficient conditions for guaranteeing a path in the flat output is exactly followable by a constrained differentially flat model are given. Consider a regular parametrized path, $P = \{ \pmb{\zeta}_{ref} \in \mathbb{R}^m  | \quad \pmb{\zeta}_{ref} = \mathbf{p}(\theta(t)), \quad \theta \in [\theta_0, \theta_1] \}$, described in the flat output space. 

\begin{theorem}

\normalfont

From \cite{c20}: Given a nonlinear \textit{differentially flat} system (1) and a path $P$ provided that $\mathbf{p}(\theta) \in C^\rho$, i.e., it is $\rho$-times differentiable and the $\rho^{th}$ derivative is continuous, where $\rho = \max{\rho_i}$, and $\mathbf{x_0} = \Phi(\cdot)|_{p(\theta(0))}$, i.e., it starts on the path and at the beginning of the path, then the dynamics of (1) under the feedforward control $\mathbf{u} = \Psi^{-1}(\cdot)|_{p(\theta(t))}$ are equivalent to a linear single-input system in normal form:
%\footnote{$\Psi^{-1}(\cdot)|_{p(\theta(t))} = \Psi^{-1}(p(\theta(t)), \frac{d p(\theta(t))}{dt}, \hdots, \frac{d^{\rho} p(\theta(t))}{dt^{\rho}})$ in (5)} 

\vspace{-3mm}
$$ \mathbf{\dot{s}} = \mathbf{A_p} \mathbf{s} + \mathbf{B_p} w, \eqno{(11)} $$

\noindent
where $ \mathbf{s} = [\theta, \dot{\theta}, \hdots, \theta^{(\rho-1)}]^T $, $ \mathbf{B_p}=[0, \hdots, 0, 1]^T$, and $\mathbf{A_p}$ contains $I^{(\rho-1) \times (\rho-1)}$ in the upper right hand corner.
	
\end{theorem}

Theorem 2 reduces the dynamics of a path-attached differentially flat system to a linear single-input model. We use this to represent the dynamics of our path-attached virtual vehicle, thereby giving it the same dynamic model that our differentially flat system exactly following the path would have. 

\section{RELATED WORK}

%\subsection{Feedforward Linearization and Prediction}
\label{sect_ff_lin_pred}

\begin{figure}[thpb]
      \centering
      \vspace{3mm}
      \includegraphics[scale=0.9]{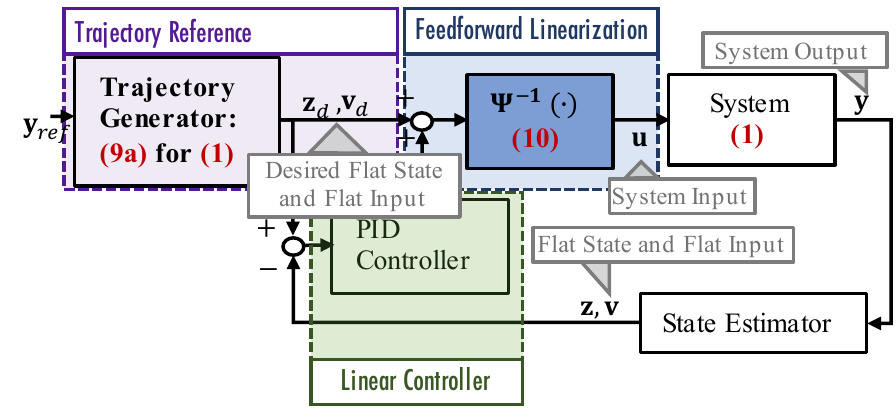}
      \caption{Early works \cite{c21}, \cite{c22} using the linear flat model (9a) for trajectory generation coupled with feedforward linearization.}
      \label{fbpc}
   \end{figure}

\begin{figure}[thpb]
      \centering
      \includegraphics[scale=0.9]{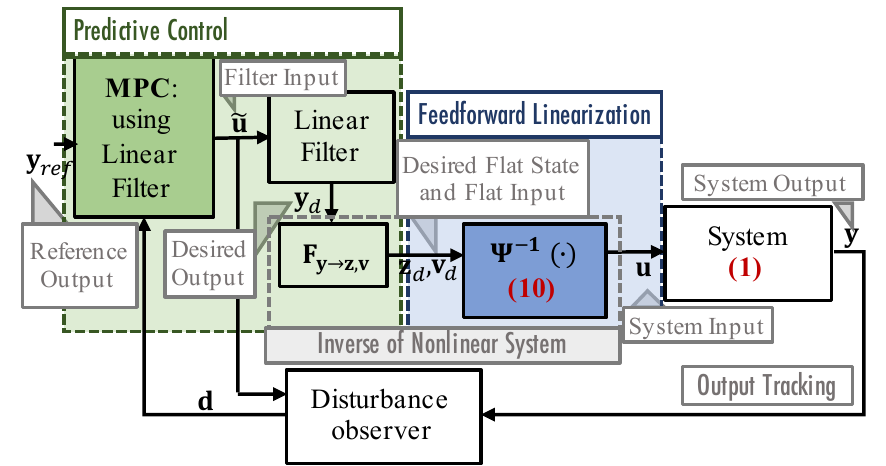}
      \caption{Prediction using a linear filter model and inversion (including feedforward linearization) for output tracking  \cite{c23}.}
      \label{fbpc2}
   \end{figure}

Feedforward linearization has been coupled with prediction or path-following before, but not with both. We showed a path-following problem that included feedforward linearization in Section \ref{sect_ff_lin_pf}. We now briefly describe related work that couples feedforward linearization and the result from Theorem 1 with prediction. Flatness-Based Predictive Control (FBPC), suggested in \cite{c21} and further developed in \cite{c22} for trajectory generation, attempts to couple prediction with feedforward linearization. As highlighted in Fig. \ref{fbpc}, the work in \cite{c21} and \cite{c22} considers using the linear flat model (9a) for trajectory generation and then combines the feedforward term (10) with a PID controller. In \cite{c15}, \cite{c5}, the differential flatness of the standard quadrotor model is utilized to generate minimum snap and jerk trajectories. In \cite{c23}, see Fig. \ref{fbpc2}, the authors use the inverse of the nonlinear model (from input $\mathbf{u}$ to output $\mathbf{y}$), a combination of the feedforward term (10) and a term $\mathbf{F_{y \rightarrow z,v}} $ that maps the output $\mathbf{y}$ to the flat state and flat input. They then only consider a linear filter model in the MPC. A simple simulation for a SISO flat model demonstrates the potential computational benefit of the combined MPC and feedforward linearization over NMPC. We use a similar idea in Section~\ref{sect_meth} to reduce the computational burden of MPC in predictive path-following for nonlinear systems.  

\section{METHODOLOGY}

\label{sect_meth}

\subsection{Overall Architecture}

We term our presented method Flatness Approach to Predictive Path-following (FAPP) because the differential flatness of the system model is pertinent to the proposed architectural design in Fig. \ref{frapp}. We propose this design for two main reasons. Unlike related works in Fig. 2 and Fig. 3, our novel combination of feedforward linearization and MPC can easily be extended to include a path dynamic for path-following. Within this design, we can significantly simplify the problem statement in Section \ref{sect_problem} by utilizing the result from Theorem 2. The proposed architecture is comprised of three key components (see Sections \ref{sect_fflin} - \ref{sect_pred}):

\textbf{Feedforward Linearization}: Utilizing feedforward linearization of the differentially flat system (1), we reduce our model to an equivalent linear flat model (9a). We use this linear flat model in our MPC. We feed the output of the MPC through an inverse term (10). 

\textbf{Path-following}: We consider a path-attached virtual vehicle, with an associated path dynamic model, $\mathbf{\dot{s}}(t) = g(\mathbf{s}(t), w(t))$ with $t \in \mathbb{R}$, the path-attached virtual vehicle state $ \mathbf{s}(t) \in \mathbb{R}^\rho$, the path-attached virtual vehicle input $ w(t) \in \mathbb{R}$ and $g$ a smooth function, attached to a parametrized geometric path in the flat output space:

\vspace{-4mm}
$$P = \{ \pmb{\zeta}_{ref} \in \mathbb{R}^m  | \quad \pmb{\zeta}_{ref} = p(\theta(t)), \quad \theta \in [\theta_0, \theta_1] \}. \eqno{(12)}$$  
\noindent 
We use Theorem 2 for differentially flat system (1), to reduce this path dynamic model to a linear single-input model (11). 

\textbf{Predictive Control}: To solve the path-following problem a sampled data MPC strategy is considered. At each sampling time, we solve an open-loop OCP, by minimizing the cost function: $J(\mathbf{\hat{x}}, \mathbf{\hat{u}}, \mathbf{\hat{s}}, \mathbf{\hat{w}}), $ which is dependent on both the sequence of predicted system states $\mathbf{\hat{x}}$ and inputs $\mathbf{\hat{u}}$ and the sequence of predicted path-attached vehicle states $\mathbf{\hat{s}}$ and inputs $\mathbf{\hat{w}}$.  This is subject to both the linear flat model of the system (9a) and the linear path dynamic model (11). We also include additional linear constraints. To ensure the system stays close to the defined geometric path and is robust to disturbances, the cost function $J(\cdot)$ weights the error between the system and path dynamic with an additional dynamic specification, such as a speed assignment. In other words, it tries to enforce that the system behaves as closely as it can to the path-attached vehicle while secondarily trying to achieve a particular dynamic specification/speed. We show that for a straight line the resulting optimization reduces to a QP. For more complex curves, we propose using one iteration of a Gauss-Newton approach to obtain a QP specifically noting that the nonlinearity is only due to the nonlinear curve. For detailed assumptions necessary for a stability proof, please see \cite{c11}.

%\displaystyle \min_{\mathbf{\hat{u}}, \mathbf{\hat{w}}} 

\subsection{Feedforward Linearization}
\label{sect_fflin}
 
The proposed coupling of feedforward linearization and MPC, as seen in Fig. 1, uses the linear flat model (9a) in a feedback MPC. The MPC outputs $\mathbf{z}_d$ and $\mathbf{v}_d$ which are then fed through the inverse term (10). Note that in this approach we ensure adherence to the initial condition requirement of feedforward linearization in Theorem 1 by feeding back our measured flat state $\mathbf{z}$ into the MPC where we re-optimize for our updated desired trajectory, $\mathbf{z}_d$ and $\mathbf{v}_d$.

\subsection{Path Following}
\label{sect_pf}

We define our path (2) in Problem 1 in the flat output space as in (12) and address geometric convergence by minimizing the difference between our system flat output and a path-attached virtual vehicle's flat output. This can be done because from Ò(4)Ó and Ò(5)Ó, the state $\mathbf{x}$ and input $\mathbf{u}$ of the nonlinear system (1) can be written directly in terms of the flat output and a finite number of its derivatives. As such defining a trajectory in the flat output $\pmb{\zeta}$ corresponds directly to an associated trajectory in $\mathbf{x}$ and $\mathbf{u}$. We now demonstrate how the reference flat state, $\mathbf{z}_{ref}$ in Fig. 1, attached to a geometric path can be simultaneously determined online with the vehicle's next desired flat state $\mathbf{z}_d$ and flat input $\mathbf{v}_d$. 

\textbf{Path Dynamics:} Considering Theorem 2, we use (11) to represent the dynamics of a virtual vehicle attached to the path. We do this because we want our virtual vehicle to be representative of the actual vehicle's motion capabilities, including necessary constraints, while being constrained to the defined geometric path. In (11), we term $\mathbf{s}$ the \textit{path state} and $w$ the \textit{path input}.

\textbf{Path Parametrization:} Although any suitable path parametrization can be used, we consider B\'{e}zier curves,
\vspace{-2mm}  
$$
\pmb{\zeta}_{ref} = p(\theta) = \sum_{j=0}^q \binom{q}{j} (1 - \theta)^{q-j} \theta^j \mathbf{P_j}, \quad \theta \in [0, 1], \eqno{(13)}
$$
\noindent
for their smoothness properties \cite{c14}.  Differentiation of (13) allows us to obtain a reference flat state $\mathbf{z}_{ref}$ parametrized in terms of the path state $\mathbf{s}$ and path parameters $\{\mathbf{P_j}\}_{j=1,\hdots,q}$:
\vspace{-3mm} 
$$
\mathbf{z}_{ref} = h(\mathbf{s}, \{\mathbf{P_j}\}_{j=1,\hdots,q}). \eqno{(14)}
$$
\noindent
For a straight line, i.e., $q=1$, (14) reduces to:
\vspace{-1mm} 
$$
\mathbf{z}_{ref} = \pmb{\Pi} \mathbf{s} + \pmb{\Pi_0}, \eqno{(15)}
$$
\noindent
where $\pmb{\Pi}$ and $\pmb{\Pi_0}$ are constant matrices containing a combination of path parameters $\mathbf{P_0}$ and $\mathbf{P_1}$. 

\subsection{Predictive Control}
\label{sect_pred}

Using the results of feedforward linearization from Sections \ref{sect_fflin} and \ref{sect_pf}, we can formulate a simplified OCP:

\begin{problem}
	\normalfont
	Given a system that can be represented by a \textit{differentially flat} nonlinear model (1) and a geometric path to be followed (12), design the flat input $\mathbf{v}(t)$ and path-attached virtual vehicle input $w(t)$ such that the following is satisfied: At every time step k, solve the OCP:

\vspace{-2mm}
$$
\min_{\mathbf{\hat{v}}, \mathbf{\hat{w}}} J(\mathbf{\hat{z}}, \mathbf{\hat{v}}, \mathbf{\hat{s}}, \mathbf{\hat{w}}), \eqno{(16)}
$$

\noindent
where we consider the sequence of predicted flat states $\mathbf{\hat{z}}$, flat inputs $\mathbf{\hat{v}}$, path states $\mathbf{\hat{s}}$ and path inputs $\mathbf{\hat{w}}$. This OCP is subject to the equivalent \textit{linear flat model} in (9a), the \textit{linear path dynamic model} in (11) (which gives a corresponding reference flat state, $\mathbf{z}_{ref}$, through path parameterization (14)) and linear constraints on the optimization variables (the flat inputs and path inputs). 

\end{problem}

We propose an objective function, $J(\cdot)$ in (16), of the form:

\vspace{-4mm}
$$ \frac{1}{2} \sum_{k=1}^N [ \mathbf{e}^{\mathbf{T}}_{p,k} \mathbf{\tilde{Q}} \mathbf{e}_{p,k} + \mathbf{e}^{\mathbf{T}}_{v,k} \mathbf{\tilde{S}} \mathbf{e}_{v,k} + \mathbf{v}^{\mathbf{T}}_k \mathbf{\tilde{R}} \mathbf{v}_k + \tilde{R}_p w^2_k ],
\eqno{(17)}
$$

\noindent
where the first term weights the positional error $\mathbf{e}_{p,k}$, at timestep $k$, between the vehicle's and reference's flat outputs (geometric convergence in Problem 1):

\vspace{-3mm}
$$
\mathbf{e}_{p,k} = \pmb{\zeta}_{k} - \pmb{\zeta}_{ref, k}, 
$$

\noindent
while the second term tries to propel the vehicle forward with a certain velocity (dynamic specification in Problem 1):

\vspace{-3mm}
$$
\mathbf{e}_{v,k} = \pmb{\dot{\zeta}}_{k} - \pmb{\dot{\zeta}}_{cmd, k}, 
$$

\noindent
where $\pmb{\dot{\zeta}}_{cmd,k} = (\frac{d \pmb{\zeta}_{ref}}{dt}|_{\dot{\theta}=v_{cmd}})_k$ in (13) and $\mathbf{\tilde{Q}}$, $\mathbf{\tilde{S}}$, $\mathbf{\tilde{R}}$ are positive semi-definite. Note $v_{cmd}$ is some designed desired speed. Notice that for the straight line case $\pmb{\dot{\zeta}}_{cmd,k} = (\mathbf{P_1} - \mathbf{P_0}) v_{cmd}$ is a constant. The last two terms in the objective function ensure regularization of the inputs. 

%\textbf{P3 (Constraint Satisfaction)} Implement MPC, $\forall t \in [t_k, t_{k+N}]$, adhering to linear constraints on the optimization variables (the flat inputs and path inputs):

%$$
%\mathbf{G} \mathbf{v} \leq \mathbf{h} \eqno{(17a)}
%$$

%\vspace{-3mm}
%$$
%\mathbf{G_p} w \leq \mathbf{h_p} \eqno{(17b)}
%$$
%\end{problem} 

As mentioned, the proposed FAPP scheme is an extended MPC scheme. The open-loop OCP is expanded by the virtual state $\mathbf{s}$ and by the virtual input $w$. Essentially, $w$ controls the path parameter evolution. Considering the FAPP scheme, path convergence can be weighted to be more important than speed. Note that preceding work on NMPC for path following, \cite{c12} and \cite{c14}, placed the velocity command on the virtual vehicle instead of on the actual vehicle. When the velocity command is placed on the virtual vehicle, in the event that the actual vehicle is disturbed, the virtual vehicle continues to try and reach the command velocity while pushing the actual vehicle to its saturation limits. Instead, we place the velocity command on the actual vehicle such that in the event of such a disturbance, the virtual vehicle tries to remain as close to the actual vehicle as possible, thus allowing better recovery from disturbances and mitigating potential saturation. 

We can demonstrate that following a straight line, parametrized as a linear B\'{e}zier curve, reduces the OCP (16)-(17) to a QP. By using that $\pmb{\zeta}_{k}$ is a component in $\mathbf{z}_k$, we can rewrite the first term in (17):
$\mathbf{e}^{\mathbf{T}}_{p,k} \mathbf{\tilde{Q}} \mathbf{e}_{p,k} = (\mathbf{z}_k - \mathbf{z}_{ref,k})^{\mathbf{T} }\mathbf{Q} (\mathbf{z}_k - \mathbf{z}_{ref,k}) 
$ where $ \mathbf{Q} \in \mathbb{R}^{\bar{\rho} \times \bar{\rho}} $, $\bar{\rho} := \sum_{i=1}^m \rho_i$, is a diagonal matrix with $\mathbf{Q}_{1,1} = \mathbf{\tilde{Q}}_{1,1}$ and $\mathbf{Q}_{\rho_i+1, \rho_i+1} = \mathbf{\tilde{Q}}_{i+1,i+1}$, $i=1,\hdots,m-1$. We similarly rewrite the second term in (17):
$
\mathbf{e}^{\mathbf{T}}_{v,k} \mathbf{\tilde{S}} \mathbf{e}_{v,k}  = (\mathbf{z}_k - \mathbf{z}_{cmd,k})^{\mathbf{T}} \mathbf{S} (\mathbf{z}_k - \mathbf{z}_{cmd,k}) 
$ where $ \mathbf{S} \in \mathbb{R}^{\bar{\rho} \times \bar{\rho}} $ is a diagonal matrix with $\mathbf{S}_{2,2} = \mathbf{\tilde{S}}_{1,1}$ and $\mathbf{S}_{\rho_i+2, \rho_i+2} = \mathbf{\tilde{S}}_{i+1,i+1}$, $i=1,\hdots,m-1$, and $\mathbf{z}_{cmd, k} = \mathbf{z}_{ref, k}|_{\dot{\theta}=v_{cmd}}$ in (14).  We then discretize the linear flat model (9a):
\vspace{-2mm}
$$
	\mathbf{z}_{k+1} = \mathbf{A_d} \mathbf{z}_k + \mathbf{B_d} \mathbf{v}_k.
$$

\noindent
We also discretize the linear path model (11):
\vspace{-2mm}
$$
		\mathbf{s}_{k+1} = \mathbf{A_{pd}} \mathbf{s}_k + \mathbf{B_{pd}} {w}_k.
$$

\noindent 
Given a current measured flat state, $\mathbf{z}_0$, and a current path state, $\mathbf{s}_0$, we write lifted forms, for $N$ prediction steps, of our discretized models:

\vspace{-5mm}
$$
\small
 \begin{aligned}
 	 	\underbrace{\begin{bmatrix}
 		\mathbf{z}_1 \\
 		\mathbf{z}_2 \\
 		\vdots \\
 		\mathbf{z}_N
 	\end{bmatrix}}_{\mathbf{\hat{z}}} = \underbrace{\begin{bmatrix}
 		\mathbf{A_d} \\
 		\mathbf{A^2_d} \\
 		\vdots \\
 		\mathbf{A^N_d} \\
 	\end{bmatrix}}_{\mathbf{\hat{A}}} \mathbf{z}_0 + \underbrace{\begin{bmatrix}
 		\mathbf{B_d} \quad 0 \quad \qquad 0 \hdots 0 \\
 		\mathbf{A_d}\mathbf{B_d} \mathbf{B_d} \qquad 0 \hdots 0 \\
 		\vdots \qquad \ddots \qquad \ddots \qquad \\
 		\mathbf{A^{N-1}_d}\mathbf{B_d} \hdots \mathbf{A_d}\mathbf{B_d} \mathbf{B_d} 
 	\end{bmatrix}}_{\mathbf{\hat{B}}} \underbrace{\begin{bmatrix}
 		\mathbf{v}_1 \\
 		\mathbf{v}_2 \\
 		\vdots \\
 		\mathbf{v}_N \\
 	\end{bmatrix}}_{\mathbf{\hat{v}}} 
 \end{aligned}. \eqno{(18)}
$$ 

\noindent
And in a similar manner, 

\vspace{-3mm}
$$
	 	\mathbf{\hat{s}} = \mathbf{\hat{A}_p} \mathbf{s}_0 +  \mathbf{\hat{B}_p} \mathbf{\hat{w}}.
$$

\noindent
We also define the expanded forms of our weight matrices: $ \mathbf{\hat{Q}} \in \mathbb{R}^{N\bar{\rho} \times N\bar{\rho}}$ where $ \mathbf{\hat{Q}} = diag(\mathbf{Q})$ and similarly for $\mathbf{\hat{S}}$, $\mathbf{\hat{R}}$ and $\mathbf{\hat{R}_p}$. Further, we define: $\mathbf{\hat{\Pi}} = diag(\mathbf{\Pi})$, $ \mathbf{\hat{\Pi}_0} = [\mathbf{\Pi_0}, \mathbf{\Pi_0}, \hdots, \mathbf{\Pi_0}]^T$ and $ \mathbf{\hat{z}}_{cmd} = [\mathbf{z}_{cmd,1}, \mathbf{z}_{cmd,2}, \mathbf{\hdots, z}_{cmd,N}]^T$. We rewrite (17) using: $\sum_{k=1}^N \mathbf{e}^{\mathbf{T}}_{p,k} \mathbf{\tilde{Q}} \mathbf{e}_{p,k} = (\mathbf{\hat{z}} - (\mathbf{\hat{\Pi}} \mathbf{\hat{s}} + \mathbf{\hat{\Pi}_0}) )^{T} \mathbf{\hat{Q}} (\mathbf{\hat{z}} - (\mathbf{\hat{\Pi}} \mathbf{\hat{s}} + \mathbf{\hat{\Pi}_0}))$; $\sum_{k=1}^N \mathbf{e}^{\mathbf{T}}_{v,k} \mathbf{\tilde{S}} \mathbf{e}_{v,k} = (\mathbf{\hat{z}} - \mathbf{\hat{z}}_{cmd} )^{T} \mathbf{\hat{S}} (\mathbf{\hat{z}} - \mathbf{\hat{z}}_{cmd})$; $\sum_{k=1}^N \mathbf{v}_k^T \mathbf{\tilde{R}} \mathbf{v}_k = \mathbf{\hat{v}}^{T} \mathbf{\hat{R}} \mathbf{\hat{v}}$; $\sum_{k=1}^N \tilde{R}_p w^2_k = \mathbf{\hat{w}}^{T} \mathbf{\hat{R}_p} \mathbf{\hat{w}}$.
 
Substituting in our expanded models for $\mathbf{\hat{z}}$ and $\mathbf{\hat{s}}$, we can simplify our objective function to a quadratic form $ J(\mathbf{\tilde{v}}) = \frac{1}{2} \mathbf{\tilde{v}}^{T} \mathbf{H} \mathbf{\tilde{v}} + \mathbf{f}^{T} \mathbf{\tilde{v}} $ where $\mathbf{\tilde{v}} = \begin{bmatrix} \mathbf{\hat{v}}^T & \mathbf{\hat{w}}^T \end{bmatrix}^T $ subject to linear optimization constraints.

For higher-order B\'{e}zier curves, we initialize $\mathbf{\tilde{v}} = \mathbf{\tilde{v}_0}$ and consider one iteration of a Gauss-Newton approach to solve the nonlinear least squares problem. The Gauss-Newton step fits a quadratic to the curve locally and reduces the problem to a QP at each step. This approach has been used in NMPC \cite{c14}, \cite{c9}. Notice that in our case, however, the nonlinearity is only as a result of a nonlinear curve while all models used are linear. 

\section{APPLICATION TO QUADROTORS}
\label{sect_quad}

\subsection{Model}

We describe how the presented methodology in Section~\ref{sect_meth} can be applied to a quadrotor. We refer to the \textit{standard quadrotor model}, $\mathbf{\dot{x}} = f(\mathbf{x}, \mathbf{u})$, \cite{c15}, \cite{c5} with state $ \mathbf{x} = (x, y, z, \dot{x}, \dot{y}, \dot{z}, \mathbf{R}, p, q, r)$ and input $\mathbf{u} = (u_1, u_2, u_3, u_4) = (T, \tau_{\phi}, \tau_{\theta}, \tau_{\psi})$ where $x,y,z$ are the linear position; $\dot{x},\dot{y},\dot{z}$ are the linear velocity; $\mathbf{R}$ is the rotation of the quadrotor body frame with respect to an inertial frame; $p,q,r$ are the body frame rotation rates, $T$ is the body thrust and $\tau_{\phi}, \tau_{\theta}, \tau_{\psi}$ are the respective body torques. We also include a standard model of the dynamics of an inner loop controller (tuned by control parameter $\tau$) which takes inputs $\mathbf{u}_{cmd} =(\dot{z}_{cmd}, \phi_{cmd}, {\theta}_{cmd}, r_{cmd})$, where $\dot{z}_{cmd}$ is a commanded velocity in z, $\phi_{cmd}$ and ${\theta}_{cmd}$ are commanded roll and pitch angles and $r_{cmd}$ is a commanded yaw velocity in the body frame, and outputs $\mathbf{u}$. The detailed equations are found in the supplementary material \cite{c24}.

\subsection{Differential Flatness}
\label{sect_quadflat}

The differential flatness of the quadrotor model is demonstrated in \cite{c15} for flat outputs $\pmb{\zeta} = (x, y, z, \psi)$. The flat state (6) and flat input (8) are shown to be $ \mathbf{z} = (x, \dot{x}, \ddot{x}, \dddot{x}, y, \dot{y}, \ddot{y}, \dddot{y}, z, \dot{z}, \ddot{z}, \dddot{z}, \psi, \dot{\psi})$ and $ \mathbf{v}= (x^{(4)}, y^{(4)}, z^{(4)}, \ddot{\psi})$ respectively.
 
Detailed mathematical derivations are necessary to show that the quadrotor with inner loop control dynamics does not change the differential flatness property of the original nonlinear quadrotor model (see supplementary material \cite{c24} for details). We can show that we have the same resulting linear flat model (9a) as \cite{c15}, however, the inverse term (10) changes to account for the inner loop controller. We highlight the high-level approach in derivation below. To show differential flatness we need to satisfy condition (3)-(5).
 
 First, notice that given that the flat outputs $\pmb{\zeta}$ are comprised of some terms of state $\mathbf{x}$, condition (3) is shown by definition. Similarly, condition (4) holds for the translational states of $\mathbf{x}$ by definition of the flat outputs. We are then left to derive $\mathbf{R}$ and $(p,q,r)$in terms of the flat state. We begin by writing $\mathbf{R}$ in terms of its column vectors and considering the translational acceleration in the standard quadrotor model. We derive $p,q,r$ in terms of the flat state by taking the derivative of acceleration and using our results for $\mathbf{R}$. We have shown that condition (4) holds. 
 
Finally, we demonstrate condition (5) by showing $\mathbf{u}_{cmd}$ as a function of the flat outputs and their derivatives (see \cite{c24} for details).

\subsection{Path Dynamics and Constraints}

Considering $\rho=4$ (obtained from the flatness derivation in Section \ref{sect_quadflat}), the linear path dynamic model for a virtual quadrotor vehicle is given by (11) with $\mathbf{s} = (\theta, \dot{\theta}, \ddot{\theta}, \dddot{\theta})$.

%\subsection{Incorporation of Constraints}

We consider two sets of constraints on the quadrotor: the first is a constraint on the body rates and the second is a constraint on the total thrust. We consider these constraints because they establish a fairly representative set of feasible quadrotor maneuvers while being bounded by convex
	functions, thereby still allowing a convex optimization problem. 

As in \cite{c5}, we consider the constraint on the body rates as a constraint on the jerk of the quadrotor: $\mathbf{j_{min}} \leq (\dddot{x}, \dddot{y},\dddot{z}) \leq \mathbf{j_{max}}$ where $\mathbf{j_{min}}$ and $\mathbf{j_{max}}$ are the minimum and maximum jerk, respectively. Given that this jerk constraint is affine in the flat state and we have a linear flat model (9a), the resulting constraint is linear in the optimization variables $\mathbf{\tilde{v}}$. 
 
 Our second constraint on the maximum thrust \cite{c5} is given as: $\ddot{x}^2 + \ddot{y}^2 + (\ddot{z}+g)^2 \leq f^2_{max}$ where $f_{max}$ is the maximum total thrust $T$ that the quadrotor can produce. The discretized version of the constraint can be put in lifted form (18) resulting in an inequality that is quadratic in $\mathbf{\hat{z}}$. We make the assumption that the quadratic coefficient is small based on the fact that it contains squares of relatively small values in $\mathbf{\hat{B}}$, obtained through discretization of the linear flat model (9a). We, therefore, reduce the maximum thrust constraint to a linear constraint on the optimization variables $\mathbf{\tilde{v}}$. 
 
 %Defining $\mathbf{\tilde{G}} \in \mathbb{R}^{3N}$ where $\mathbf{\tilde{G}}_{3k}=g \quad k=1,\hdots,N$, $\mathbf{\tilde{M}} \in \mathbb{R}^{3N \times 14N}$ where $\mathbf{\tilde{M}}_{3k+1,14k+3} = 1, \quad \mathbf{\tilde{M}}_{3k+2,14k+7} = 1, \quad \mathbf{\tilde{M}}_{3k+3,14k+11} = 1 \quad k=0,\hdots,N-1$ and $\mathbf{F} \in \mathbb{R}^{3N}$ where $ \mathbf{F}_{k} = f^2_{max} \quad k=1,\hdots,N$ we can rewrite our maximum thrust constraint as: $
 %(\mathbf{\tilde{M}} \mathbf{\hat{z}} + \mathbf{\tilde{G}})^T (\mathbf{\tilde{M}} \mathbf{\hat{z}} + \mathbf{\tilde{G}}) < \mathbf{F}$. Plugging in our expanded discretize model for $\mathbf{\hat{z}}$ and making the assumption that  $\mathbf{\hat{B}}^T \mathbf{\tilde{M}}^T \mathbf{\tilde{M}} \mathbf{\hat{B}}$ is relatively small, we can reduce the maximum thrust constraint to a linear constraint on the optimization variables $\mathbf{\tilde{v}}$. 

% $\mathbf{A_{con}} \mathbf{v} \leq \mathbf{B_{con}}$ where 

%$$
%	\mathbf{ A_{con} } = 2 \mathbf{z_0}^T \mathbf{\hat{A}}^T \mathbf{\tilde{M}}^T \mathbf{\tilde{M}} \mathbf{\hat{B}} + 2 \mathbf{G}^T \mathbf{\tilde{M}} \mathbf{\hat{B}}
%$$

%\vspace{-5mm}
%$$
%	\mathbf{B_{con}} = \mathbf{F} - \mathbf{G}^T \mathbf{G} - 2 \mathbf{G}^T \mathbf{\tilde{M}} \mathbf{\hat{A}} \mathbf{z_0} - \mathbf{z_0}^T \mathbf{\hat{A}}^T \mathbf{\tilde{M}}^T \mathbf{\tilde{M}} \mathbf{\hat{A}} \mathbf{z_0}
%$$

\section{EXPERIMENT}
\subsection{Setup}

The experiments are conducted on a Parrot AR.Drone quadrotor with an overhead motion capture system estimating the state of the quadrotor. We interface with the quadrotor using the open-source Robot Operating System (ROS). The state estimator in Fig. 1 determines the flat state $\mathbf{z}$ at 200~Hz. 

%The state estimate, comprising of state $\mathbf{x}$ in the standard quadrotor model, is received at 200 Hz.

 %The inner loop is modelled as (18a)-(18k) with parameters $\tau_{Iz} = 0.05$, $\tau_{I\psi} = 0.01$, $\tau_{rp} = 0.18$, $\tau_{r} = 0.1$, $\tau_{p}=0.18$, $\tau_{q}=0.18$. 

The control architecture consists of two loops: an off-board outer loop running at 70 Hz and an on-board inner loop running at 200 Hz. In the experiments we compare two outer loop controllers: the first uses our proposed FAPP architecture, see Fig. 1, and the second uses a baseline nonlinear trajectory tracking controller.  The baseline nonlinear trajectory tracking controller computes the desired $x$ and $y$ accelerations using PD control and then determines the associated pitch $\theta_{cmd}$ and roll $\phi_{cmd}$ commands using standard feedback linearization. The z-velocity $\dot{z}_{cmd}$ and yaw-velocity $\dot{\psi}_{cmd}$ commands are computed using P control. In the FAPP controller, the MPC considers a look ahead time of $1 s$ where the prediction horizon is $N=10$. We consider the following weights in our objective function (16): $\mathbf{\tilde{Q}} = diag(500)$, $\mathbf{\tilde{S}} = diag(10)$, $\mathbf{\tilde{R}} = diag(0.01)$, $\tilde{R}_p = 0.1$. In terms of constraints, we consider $\mathbf{j_{min}} = [-20 -20 -20]^T \quad m/s^3$, $\mathbf{j_{max}} = [20 \quad 20 \quad 20]^T \quad m/s^3$ and $f_{max} = 20 \quad m/s^2$.

%and the linear flat model is discretized at 10 Hz

We perform the experiments on a petal shaped path (the grey geometric path in Fig. 4) parametrized in terms of a cubic B\'{e}zier curve with: $\mathbf{P_0} = [0.0, 0.0, 1.0, 0.0]$, $\mathbf{P_1} = [3.0, 2.0, 0.0, 0.0]$, $\mathbf{P_2} = [3.0, -2.0, 0.0, 0.0]$, $\mathbf{P_3} = [0.0, 0.0, 1.0, 0.0]$. We start the reference virtual vehicle on the path at $\mathbf{P_0}$ (i.e., $\theta=0$). We investigate three cases: (i)~the quadrotor starts within 0.1 m of the reference virtual vehicle and is then required to fly the geometric path (in red in Fig. 4); (ii) the quadrotor starts more than 1.0 m from the reference virtual vehicle and is then required to converge to and fly the geometric path (in blue in Fig. 4); (iii) the quadrotor starts within 0.1 m of the reference virtual vehicle but is then disturbed by wind during its flight (in green in Fig. 6). We first fly case (i) with the FAPP controller which simultaneously computes the timed reference $\mathbf{z}_{ref}$ during the flight. We then use this generated trajectory as our nominal trajectory for the baseline trajectory tracking controller. All subsequent flights, cases (i) - (iii), with the trajectory tracking controller use this nominal trajectory. 

   \begin{figure}[thpb]
      \centering
      \includegraphics[scale=1.0]{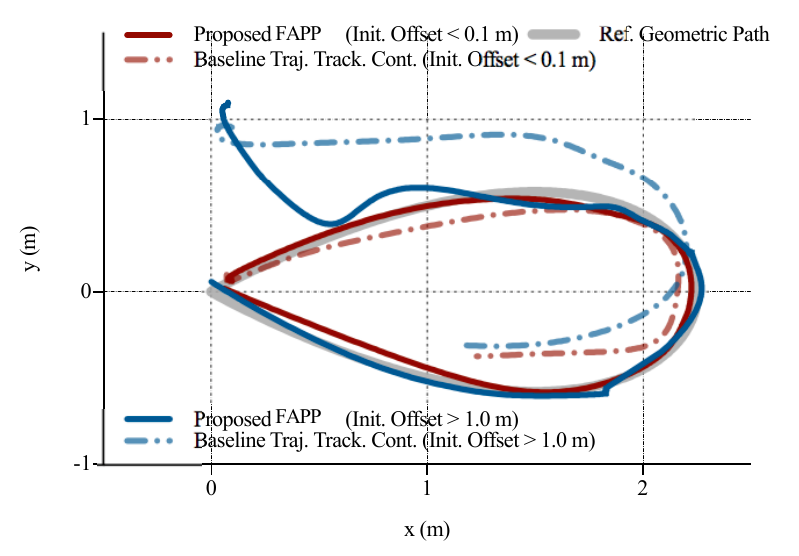}
      \caption{Comparison of proposed FAPP with baseline trajectory tracking controller. Proposed FAPP stays closer to desired geometric path (grey) in both the nominal (max. reference speed $\approx 2$ m/s) case (red) and when the initial condition is offset by $> 1.0$ m (blue).}
      \label{petalvisual}
   \end{figure}
   
      \begin{figure}[thpb]
      \centering
      \includegraphics[scale=1.0]{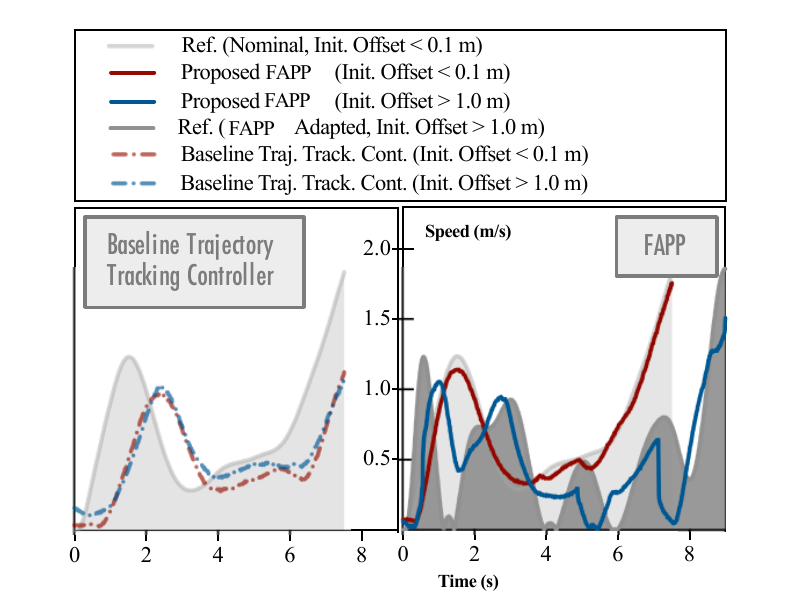}
      \caption{Speed profiles of experiment in Fig. 4. Three key observations: (1) proposed FAPP adjusts and slows down the reference speed compared to the nominal trajectory under a large initial offset (light vs. dark grey on the right); (2) in the nominal case (light grey) the proposed FAPP better tracks the reference speed (red solid vs red dotted); (3) when the initial condition is offset by $> 1.0$ m, the difference between the virtual vehicle moving on the path and the quadrotor being off the path result in the discrepency between the quadrotor and reference speeds (blue verse dark grey on right).}
      \label{frappspeed}
   \end{figure}

\subsection{Results}

Case (i): In this case, as depicted in red in Fig. 4, FAPP reduces the root mean square (RMS) error between the quadrotor position and reference position by 90.92\%. Under nominal conditions with small initial position error, the primary advantage of FAPP is that it can predict ahead, adjust the speed of the quadrotor accordingly (speeding up on straight sections of the path), and compute an input to the vehicle that accounts for upcoming turns. As demonstrated in Fig. 5, because the baseline trajectory tracking controller does not predict ahead, the quadrotor does not accelerate fast enough in the first 2 s and therefore falls behind resulting in a final position that is 1.26 m behind the reference in Fig. 4. 

   %Speed profiles of AR.Drone and reference in Cases (i)-(ii) of Vicon experiments with proposed FAPP controller and with a baseline trajectory tracking controller. Case (i): quadrotor flies at similar speed to the reference (Ref. (Nominal, Init. Offset $<$ 0.1 m)) with outer loop controller FAPP (red solid line) while it does not accelerate fast enough with the baseline trajectory tracking controller (red dotted line). Case (ii): with the FAPP outer loop controller (solid blue line), the FAPP adapted reference (Ref. (FAPP Adapted, Init. Offset $>$ 1.0 m)) changes its speed (dark grey).
\begin{figure}[thpb]
      \centering
      \includegraphics[scale=1.0]{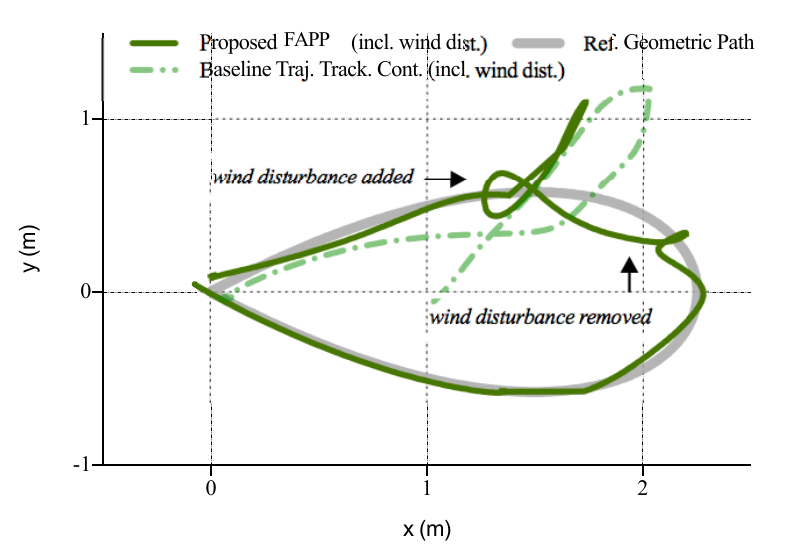}
      \caption{Comparison of proposed FAPP with baseline trajectory tracking controller under wind disturbance. Proposed FAPP stays closer to desired geometric path (grey).}
      \label{petalvisual_dist}
   \end{figure}
   
   %Visualization of AR.Drone flights in Case (iii) of Vicon experiments with proposed FAPP controller and with a baseline trajectory tracking controller. Case (iii): quadrotor stays closer to the reference geometric path (grey solid path) with outer loop controller FAPP (green solid path) than with the baseline trajectory tracking controller (green dotted path) by adapting the reference.
   
   \begin{figure}[thpb]
      \centering
      \includegraphics[scale=1.0]{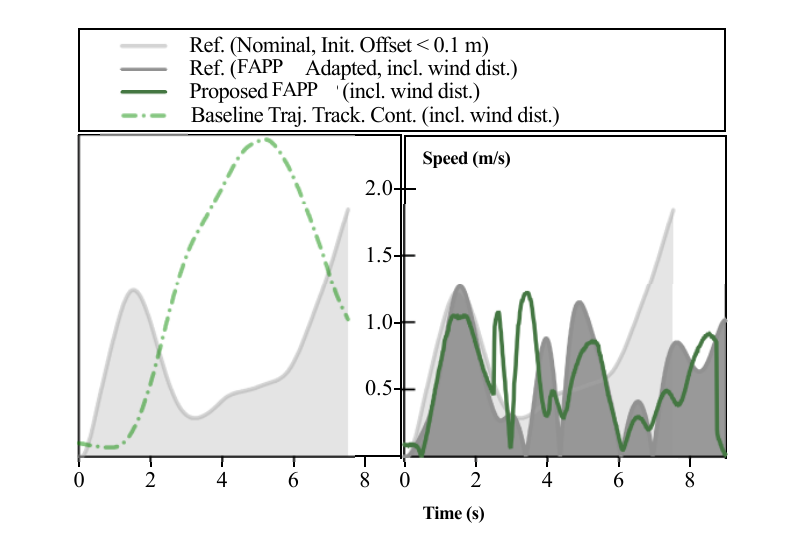}
      \caption{Speed profiles of experiment in Fig. 6. Two key observations: (1) the proposed FAPP adjusts and slows down the reference speed compared to the nominal trajectory (light vs. dark grey on the right); (2) the baseline trajectory tracking controller instead speeds up the quadrotor as it falls behind the reference (dotted green versus light grey on left).}
      \label{frappspeed_dist}
   \end{figure}

Case (ii): When the quadrotor has significant initial offset, the main advantage of FAPP as demonstrated in Fig. 4 is that it simultaneously adapts the reference trajectory in the optimization problem. As shown in Fig. 5, in the first 1 s both the quadrotor and reference speed up where the quadrotor tries to quickly converge to the path. At about 1.5~s, the quadrotor is close enough to the path but needs to change direction, correspondingly the reference waits for it to do so before both speed up once again. At about 3 s, the quadrotor begins to slow down as it makes the turn, which it completes at about 7 s. Although at about 7.2 s there is a small communication delay with the quadrotor, seen as a slight drift in Fig. 4, the quadrotor manages to quickly recover to speed up for the final segment of the path. %The reference actually slows down at about 7.2 s, allowing better quadrotor recovery.

Case (iii): We demonstrate the robustness of FAPP by adding a wind disturbance. As seen in Fig. 6 and Fig.~7, FAPP (green solid path) enhances disturbance recovery by slowing down the reference such that it waits for the disturbed quadrotor.

\addtolength{\textheight}{-7cm} 

\section{CONCLUSION}

In this paper, we presented a Flatness Approach to Predictive Path-following (FAPP), which combined feedforward linearization, model predictive control and path-following. This combination has the following advantages:

\vspace{-1mm}
\begin{itemize}
	\item Its predictive capabilities allow high-accuracy tracking around curved paths while allowing us to consider key constraints within the control design.
	\item By simultaneously adapting the path reference, without significant additional computation, we are able to enhance robustness to disturbances.
	\item We simplify the optimal control problem to solving one quadratic program by utilizing the differential flatness property to reduce both our nonlinear path and system models to equivalent linear ones. 
\end{itemize}

\vspace{-1mm}
\noindent
Furthermore, we demonstrate these advantages in experiment by implementing the presented control architecture on a quadrotor. 

%\addtolength{\textheight}{20cm}   % This command serves to balance the column lengths

%%%%%%%%%%%%%%%%%%%%%%%%%%%%%%%%%%%%%%%%%%%%%%%%%%%%%%%%%%%%%%%%%%%%%%%%%%%%%%%%

%%%%%%%%%%%%%%%%%%%%%%%%%%%%%%%%%%%%%%%%%%%%%%%%%%%%%%%%%%%%%%%%%%%%%%%%%%%%%%%%

%%%%%%%%%%%%%%%%%%%%%%%%%%%%%%%%%%%%%%%%%%%%%%%%%%%%%%%%%%%%%%%%%%%%%%%%%%%%%%%%

\end{document}